# Image Embedding of PMU Data for Deep Learning towards Transient Disturbance Classification

Yongli Zhu, Chengxi Liu, Kai Sun
Electrical Engineering and Computer Science
University of Tennessee (UTK)
Knoxville, USA
yzhu16@vols.utk.edu, cliu48@utk.edu, kaisun@utk.edu

*Abstract*—This paper presents a study on power grid disturbance classification by Deep Learning (DL). A real synchrophasor set composing of three different types of disturbance events from the Frequency Monitoring Network (FNET) is used. An image embedding technique called Gramian Angular Field is applied to transform each time series of event data to a two-dimensional image for learning. Two main DL algorithms, i.e. CNN (Convolutional Neural Network) and RNN (Recurrent Neural Network) are tested and compared with two widely used data mining tools, the Support Vector Machine and Decision Tree. The test results demonstrate the superiority of the both DL algorithms over other methods in the application of power system transient disturbance classification.

## I. INTRODUCTION

With the increasing deployment of synchrophasors, e.g. PMUs (Phasor Measurement Units) on both transmission and distribution systems of power grids in many countries, conventional power system operation and planning practice has confronted both new challenges and new opportunities. The monitoring and control of a future smart grid should be designed and implemented within the framework of "Energy Internet" [1]. The vast PMU measurements generated continuously during the daily grid operations are big data and require the most cutting-edge data mining and machine leaning techniques to analyze and digest for useful information. Those machine leaning technologies have been utilized to improve the event detection speed and system monitoring accuracy for the security of power system [2], [3].

The PMU technology has the following features: on one hand, PMUs have a high sampling rate, e.g. 30 samples per second for a 60Hz AC power system, which is much higher than that of SCADA data. The data size for a single day could be extremely huge and hence poses a challenge for effective reprocessing and utilization of the data. On the other hand, by means of GPS (Global Positioning System), PMU measurements are synchronized and integrated into the Wide Area Measurement Systems (WAMS), which in turn provide potential applications, including but not limit to: online identification of the fault location and type [4], model reduction and model verification [5], oscillation monitoring and damping control [6]-[8], early warning of instabilities [9]-[12], dynamic security and stability assessments [13], etc.

Among all the above PMU applications, the accurate classification of disturbances is critical for power system state estimation, protection and control [14]. The existing methods to classify disturbances are mostly based on time series analysis or conventional signal processing theory. For example, in [15] a Wavelet based method is presented for event detection by utilizing the PMU measurements. In [16], a combination of the Empirical Mode Decomposition and Spectral Kurtosis methods is introduced for event signal characterization based on PMU data stream. They are efficient in real-time detection. However, to carry out more advanced system analysis (e.g. system model reduction), the accuracy of current disturbance classification approaches needs further improvement.

To detect a disturbance as early as possible, it is useful to classify ambient signals. Another issue of some existing methods is that they are dependent on operators' experience, for example, classifying the disturbance type based on a pre-defined set of threshold values. In addition, during the period of disturbances, the latent system process is in fact time-varying due to the drastic changes of the power system states. Therefore, the solution provided by conventional model-based or experience-based methods is essentially a static characterization of the real system. Therefore, measurement based methods such as machine learning can be considered here. Approaches from the machine learning area have been utilized by researchers in the power industry during the last ten years [17], [18]. Deep Learning (DL) is one of the most active areas in the field of machine learning. It has brought a revolutionary impact on the computer science area. Its powerful performance is initially demonstrated in the computer vision area such as object detection, unmanned automatic driving, etc. Recently, its application has been extended into other fields, e.g. time series analysis [19], [20].

The objective of this paper is to utilize deep learning methods to identify different types of disturbances. The main contributions of this paper include: 1) an imaging-transformation based Deep Learning method is created for the power system transient disturbance classification problem; 2) the performance of the proposed method is verified against two widely used conventional data mining methods using real PMU datasets.

This work was supported by the ERC Program of the NSF and DOE (EEC-1041877).

The remaining parts are organized as follows: a real PMU-based time series dataset is described and illustrated in Section II; Section III explains a recent reported method called "GAF" (Gramian Angular Field) [21] for image embedding of the original time series. Section IV introduces the basic principles of the two major DL architectures, i.e. CNN (Convolutional Neural Network) and RNN (Recurrent Neural Network). Test results are presented in Section V with comparisons against the results from the Support Vector Machine (SVM) and Decision Tree (DT) methods. Conclusions and discussion about future work are provided in Section VI.

## II. PROBLEM DESCRIPTION

### A. FNET System

The frequency monitoring network system (FNET) [22], operated by the University of Tennessee Knoxville and Oak Ridge National Lab (ORNL), is an easy-deploy WAMS with good dynamic accuracy and low installation cost. The GPS synchronized Frequency Disturbance Recorder (FDR) can measure voltage magnitude, angle, and frequency with a high precision from 220V or 115V outlets. These measured signals are calculated at 100 ms intervals and then transmitted across the public internet to a Phasor Data Concentrator (PDC), where the measurements are synchronized, analyzed, and archived. The FNET has been deployed since 2004. More than 150 FDRs have been installed in three North American interconnections: Eastern Interconnection (EI), Western Electricity Coordinating Council (WECC), and Electric Reliability Council of Texas (ERCOT), as shown in Fig. 1.

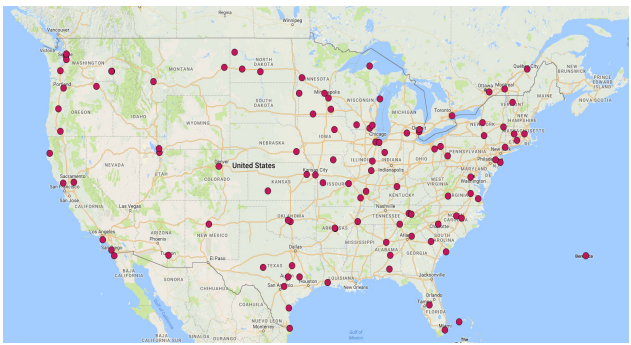

Fig. 1. FNET deployment in North America

### B. Disturbance Classification

Disturbance classification using PMU data has significant meaning for system operation, protection, control and post-event analysis. Under normal operating conditions, multiple frequency measurements across a single interconnection are identical and the variation of frequency is little. However, under disturbances, the frequency and phasor angle suffer sudden changes in a short time. It is critical to identify different types of disturbances for operators to locate the cause of an event and activate proper operations against them. In the FNET database, there are basically five types of power system disturbances: 1) Generation trip; 2) Load shedding; 3) System oscillation; 4) Line trip; 5) Islanding. In this paper, only the first three types of disturbances are available in the given period (the whole month of November 2014).

The FNET offers both frequency and angle data. In the following study, only angle data are utilized. In this study, a real dataset of 374 PMU measurements (due to the copyright agreement, the PMU station names/IDs are omitted here). The original data recording length is 1 minute with 10Hz sampling rate, thus 600 data points for each time window. For practical applications, one-min-window can be too long for pre-warning, thus in this study the first 30 sec snapshot of the PMU measurements are utilized. Among those data, 142 disturbance events are generation trips; 145 events are load-shedding; and 87 events are system oscillations. The dataset of each case includes samples from multiple sensors, each of them with a certain range of time stamp, frequency value and angle value. The dataset includes generation trips, load shedding and oscillations from the following list. Each of the case includes data from 40 to 80 FDR sensors. The plots of the three types of disturbance events are shown in Fig. 2.

TABLE I. LIST OF MEASURED DISTURBANCE EVENTS

| Disturbance type | Date / Time |
| --- | --- |
| Generation Trip | 2014/11/07 |
| Generation Trip | 2014/11/12 |
| Generation Trip | 2014/11/22 |
| Load Shedding | 2014/11/13 |
| Load Shedding | 2014/11/14 |
| Load Shedding | 2014/11/22 |
| Oscillation | 2014/11/12 |
| Oscillation | 2014/11/25 |

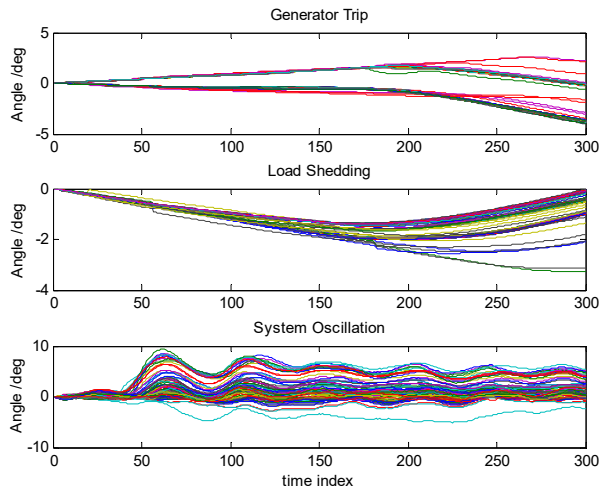

Fig. 2. PMU measurements for the three different types of disturbances

## III. IMAGE EMBEDDING OF PMU DATA

The next step is to map each PMU data (a time series) into an image, i.e. so-called "image embedding". Typically, this can be achieved by applying a certain nonlinear function. A good mapping usually satisfies the following property: 1) preserving the temporal-order information 2) the mapping is one-to-one and its inverse image exists 3) the computational burden is acceptable, as small as possible to benefit online applications. As reported in [21], the Gramian Angular Field (GAF) is a suitable choice. The math operation and definition of GAF are as follows:

$$\tilde{x}_i = \frac{x_i - \min(X)}{\max(x) - \min(X)}, \quad i = 1, 2, ..., n \quad (1)$$

$$\theta_i = \arccos(\tilde{x}_i), \tilde{x}_i \in X; \quad i = 1, 2, ..., n \quad (2)$$

$$G = \begin{bmatrix} \cos(\theta_1 + \theta_2) & \cdots & \cos(\theta_1 + \theta_n) \\ \cos(\theta_2 + \theta_1) & \cdots & \cos(\theta_2 + \theta_n) \\ \vdots & \ddots & \vdots \\ \cos(\theta_n + \theta_1) & \cdots & \cos(\theta_n + \theta_n) \end{bmatrix} \quad (3)$$

where,

$X = [x_1, x_2… x_n]$ is the time-series with length $n$; $\tilde{x}_i$ is the scaled value;

$\theta_i$ is the transformed angle values in the latent polar-coordination system;

$G$ is the final Gramian Angular Field matrix corresponding to the original time-series $X$.

The benefits of GAF are 1) the time-order increases from its top-left corner to bottom-right corner. Thus, the temporal information (correlations among the different time steps for each time series) can been preserved, in other words, it is a "*matrixify*" of the time-series 2) it is easy to calculate. No integral or convolution operations but just simple cosine functions involved. To better demonstrate this "Image Embedding" idea, three time-series samples for different types of disturbances are plotted in Fig. 3, together with their corresponding GAF-mapped images by Eq. (3).

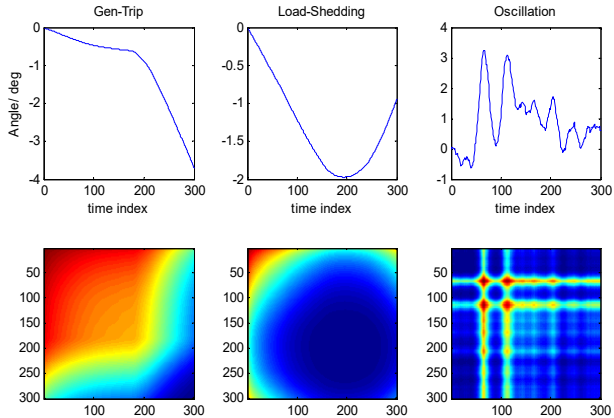

Fig. 3. Image embedding examples for PMU time series

From the mapped images, it is obvious that the differences between the three classes have been "visually augmented" and displayed more "vividly" in the embedded image space. Similar to the kernel function of SVM, the image embedding technique here introduces a "feature" augmentation" benefit, which is helpful for the classifier performance improvement. Moreover, this GAF based feature augmentation technique only involves simple triangular function and matrix arrangement operations, thus the computational cost of such data transformation is acceptable.

IV. BASIC PRINCIPLES OF DEEP LEARNING

The principles of CNN and RNN are briefly introduced in this section. As the two mainstreams in the DL area, both methods have been successfully deployed in many cutting-edge applications such as image pattern recognition and natural language processing. More details can be referred to [23], [24]. Both include a big family of various versions, here, to give a whole picture of the two structures, the most representative one for each structure will be introduced in the following paragraphs, i.e. LeNet for CNN and LSTM (Long Short-Term Memory) for RNN.

*A. CNN and LeNet introduction*

LeNet is the first successfully applied CNN structure. It takes image inputs. Two important concepts are *Convolution* and *Pooling* operations as explained below.

- Pooling/Subsampling:

Pooling is a procedure that takes input over a certain area and reduces that to a single value (subsampling), e.g., the *Max-Pooling*. It partitions the input image into a set of non-overlapping rectangles. Then, for each sub-region, it outputs the maximum value of that region. It is actually a dimension reduction process for the outputs of certain hidden layers.

- Convolutional layer

As shown in Fig. 4, it is in fact a series of linear operation on tensors/matrices to extract high-level "hidden" information.

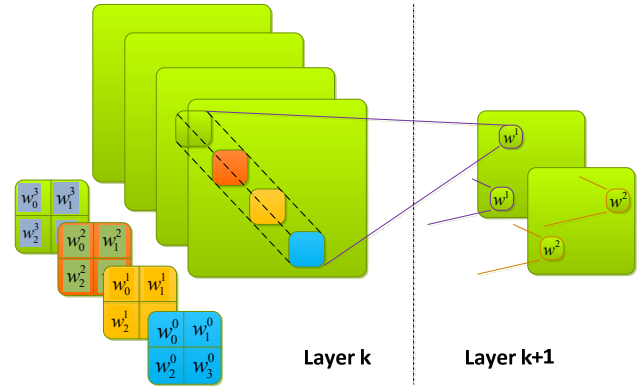

Fig. 4. Example of a convolutional layer

Finally, the LeNet structure is depicted in Fig. 5, where "C" means Convolutional layer; "S" means pooling layer; "F" means fully-connected layer (i.e. a layer with each of its neural units connected to each element of its input).

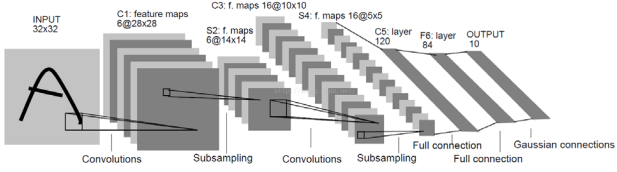

Fig. 5. The complete CNN (LeNet-5) architecture [23]

*B. RNN and LSTM introduction*

RNN adds inter-layer connections ("weights") among each hidden layer. Thus, it can be regarded as an enhanced version of traditional multi-layer neural network. It is especially suitable for learning sequential data. A basic RNN unit [24] is shown in Fig. 6.

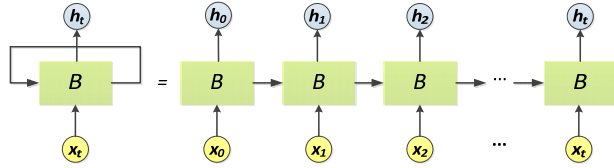

Fig. 6. A basic RNN unit and its unrolled structure

In the above diagram, a block of neural network, i.e. **B**, receives inputs $x_t$ and gives outputs $h_t$ based on a recurrent-style training approach as shown in its unrolled structure. This allows information to be passed from one step of the network to the next step.

One main drawback of traditional RNN is the "Long-Term Dependency" issue, i.e. with the time steps increasing, the gradient involved in training will be exploding or vanishing quickly. For this reason, LSTM was proposed. Different from simple RNN, LSTM tries to "remember" information within long time steps. More specifically, a basic LSTM unit has several blocks interacting in a special way as shown in Fig. 7.

The innovations of LSTM are mainly in two new concepts, "gate" ($\sigma$) and cell state ($s_t$). Gates are used to optionally allow information to pass through. Mathematically, it is simply an activation function (e.g. *sigmoid*) with pointwise multiplication operations. The gates output values between 0 and 1, describing the portion of information to be remembered. A value of 0 means "forget", while a value of 1 means "remember". An LSTM has three kinds of these gates: a forget gate $f_t$, an input gate $i_t$ and an output gate $o_t$, to control the cell states at step-$t$ as shown in Fig.7. The complete updating equations for one basic LSTM unit (neuron) used in above gates are summarized in Eq. (4).

$$\begin{cases} f_t = \sigma(W_f x_t + U_f h_{t-1} + b_f) \\ i_t = \sigma(W_i x_t + U_i h_{t-1} + b_i) \\ o_t = \sigma(W_o x_t + U_o h_{t-1} + b_o) \\ s_t = f_t \circ s_{t-1} + i_t \circ \tanh(W_s x_t + U_s h_{t-1} + b_s) \\ h_t = o_t \circ \tanh(s_t) \end{cases} \quad (4)$$

Where, "∘" stands for the inner product and:

$x_t$: input of step $t$
$s_t$: cell state of step $t$
$h_t$: output of step $t$
$\sigma$: activation function (*sigmoid*)
$W_k, U_k, b_k$: weight matrices or vectors ($k = f, i, o, s$)
$f_t, i_t, o_t$: different gate outputs of step $t$

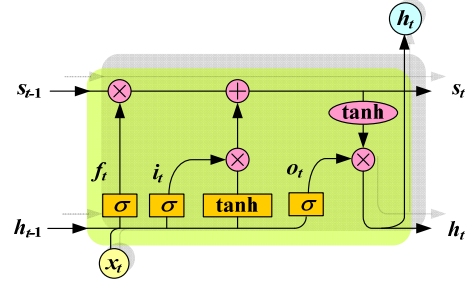

Fig. 7. The internal structure of a basic LSTM cell unit

*C. Architecture selection consideration*

From the above description of the basic principles for the two architectures, CNN is more suitable to the problem with some spatial correlation like images data; while RNN (LSTM) is more flexible in handling time series objects with temporal dynamics. However, the RNN structure can be also applied to the image data by certain data-reshaping trick. More specifically, for an image expressed in an n-by-m matrix, RNN can treat each row of that image as one "time step" of the total n steps, while each step input is an m-dimensional vector; thus, the image data here is analogous to a multivariate time series. In our study, the "many-to-one" structure RNN is chosen. It is illustrated in Fig. 8, where the input (bottom-level blocks) represents each row of the image mapped from the previous GAF method.

After all the image rows have been scanned and trained, the final output is the class information. More concretely, the final output is a one-hot encoding vector representing the probabilistic distribution of different classes for that "time series". For example, for a 3-class problem, assuming the true class of one such time series is type-3. Then, its "one-hot" encoded vector is [0, 0, 1]. The CNN or RNN output, for example, can be [0.2, 0.3, 0.5], of which the components sum to 1 and each component stands for the corresponding probability of each class.

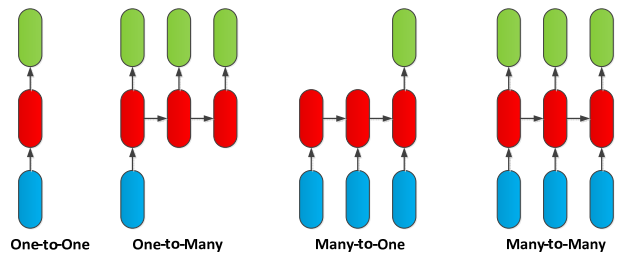

Fig. 8. Different RNN input/output paradigms

Finally, the flowcharts of the procedures using the SVM (or DT) method and Deep Learning method are shown in Fig. 9.

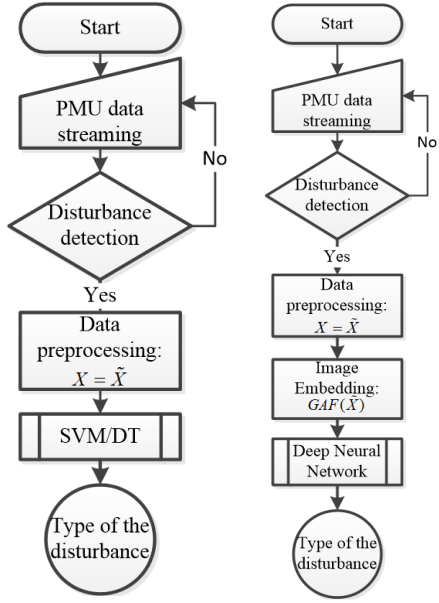

Fig. 9. Comparison of the flowcharts of the SVM/DT method (left) and the Deep Learning method (right) for disturbance classification

## V. CASE STUDY

For simplicity, three sets of experiments are designed here, i.e. using respectively 2/3, 3/4 and 4/5 of the original dataset for training, and the remaining data for testing. The tests are on a Desktop PC with Intel Core™ i7-3770 CPU (3.40GHz) and no GPU. All the experiments here are implemented in Google's *TensorFlow* and Python 3.5. Data preprocessing and post-processing are done in MATLAB. In the following tables, "Accuracy" means the metric value based on the testing dataset. Comparisons with DT and SVM are presented as well. The numerical results for each experiment are listed in Table II to Table IV. The accuracy comparison plot for all the three experiments is depicted in Fig. 10.

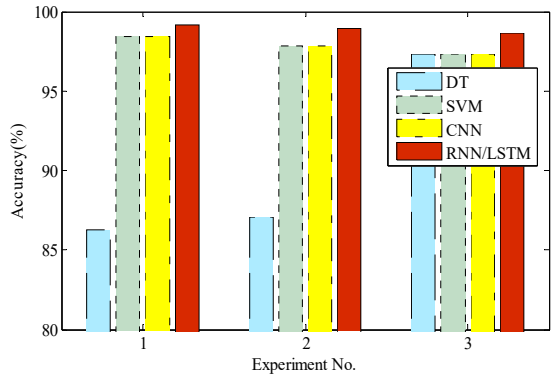

Fig. 10. The accuracy comparison plot for all the experiments

### A. CNN Results

A modified LeNet is adopted here. The only changes are 1) modification of the number of the 1st Convolutional Layer units from 6 to 32; 2) modification of the number of output dimensions from 10 to 3 for our three-type classification problem. The specific meaning of each parameters is annotated, and more details can be found in [25].

TABLE II.  CNN RESULTS

| Training parameters | values |
|---|---|
| Batch size | 64 |
| Training epochs | 30 |
| Learning rate | 0.01 |
| Momentum | 0.9 |
| Algorithm | Stochastic Gradient Descent |
| Neurons in 1st Conv. Layer | 32 |
| **Accuracy (2/3 data for train)** | **98.39%** |
| **Accuracy (3/4 data for train)** | **97.85%** |
| **Accuracy (4/5 data for train)** | **97.30%** |

### B. RNN Results

TABLE III.  RNN RESULTS

| Training parameters | RNN | LSTM |
|---|---|---|
| Hidden layer number | 1 | 1 |
| Neurons in hidden layer | 128 | 64 |
| Learning rate | 0.001 | 0.001 |
| Training epochs | 50 | 30 |
| Algorithm | Adam | Adam |
| **Accuracy (2/3 data for train)** | **99.19%** | **99.19%** |
| **Accuracy (3/4 data for train)** | **98.92%** | **98.92%** |
| **Accuracy (4/5 data for train)** | **98.65%** | **98.65%** |

### C. Comparison with other methods

The details regarding the meaning of DT and SVM parameters can be referred to [13], [18].

TABLE IV.  DT AND SVM RESULTS

| Training parameters | DT | SVM |
|---|---|---|
| Kernel functions | - | RBF |
| C, gamma | - | 1, 0.033 |
| min_samples_split | 2 | - |
| min_samples_leaf | 1 | - |
| criterion | Gini Index | - |
| **Accuracy (2/3 data for train)** | **86.29%** | **98.39%** |
| **Accuracy (3/4 data for train)** | **87.10%** | **97.85%** |
| **Accuracy (4/5 data for train)** | **97.30%** | **97.30%** |

## VI. CONCLUDING REMARKS

From the above results, the following conclusions are obtained:

*1)* Both CNN and RNN (LSTM) achieve more than 98% accuracy for this three-type disturbance classification problem.

*2)* RNN and LSTM obtain the highest accuracy (e.g. 99.19% in experiment-1) among all the classifiers. LSTM

needs less training epoch (e.g. 30) than simple RNN and CNN (i.e. 50) to reach the same level accuracy during training.

This study provides an application paradigm to connect the DL with conventional power systems studies, where the DL outperform conventional classification methods in this specific power system problem in terms of the accuracy. With the powerful classification and regression capabilities of DL techniques, it can be envisioned that a large variety of potential applications in power systems can be exploited, e.g. controller design and model identification. On the other hand, with the rapid progress of DL techniques, more advanced architectures will be utilized on power system applications, e.g. Denoise Autoencoder (DAE), Deep Boltzmann Machine (DBM), Generative Adversarial Network (GAN), etc.

ACKNOWLEDGMENT

The authors would like to present appreciations for the PMU data provided by Dr. Yilu Liu's research group at the University of Tennessee, Knoxville.